\documentclass[11pt,a4paper]{article}

\usepackage[T1]{fontenc}
\usepackage[utf8]{inputenc}
\usepackage{lmodern}
\usepackage[a4paper,margin=2.4cm]{geometry}
\usepackage{microtype}
\usepackage{cite}
\usepackage{amsmath,amssymb,amsfonts}
\usepackage{algorithm}
\usepackage{algorithmic}
\usepackage{graphicx}
\usepackage{textcomp}
\usepackage{xcolor}
\usepackage[font=small,labelfont=bf,labelsep=period]{caption}
\usepackage[
    colorlinks=true,
    citecolor=blue,
    linkcolor=blue,
    urlcolor=blue
]{hyperref}
\usepackage{orcidlink}
\def\BibTeX{{\rm B\kern-.05em{\sc i\kern-.025em b}\kern-.08em
    T\kern-.1667em\lower.7ex\hbox{E}\kern-.125emX}}

\newcommand{\paperTitle}{Creating Hybrid Interactive Narrative Systems with LLM: From Formal Description to Evaluation}
\hypersetup{
    pdftitle={\paperTitle},
    pdfauthor={Jules Clerc, Domitile Lourdeaux, Mohamed Sallak, Marc Ravaine},
    pdfkeywords={Experience Management, Functional Analysis, Interactive Narrative, Large Language Models, Narrative Generation}
}

\newif\ifieeenotice
\ieeenoticetrue   
\newcommand{\copyrightYear}{2026}
\def\vorCitation{}  
\def\vorDOI{}       

\begin{document}

\begin{center}
    {\LARGE\bfseries \paperTitle\par}
    \vspace{1.2em}
    {\large
    Jules Clerc\textsuperscript{1,2}\,\orcidlink{0009-0009-0825-4498},\quad
    Domitile Lourdeaux\textsuperscript{1}\,\orcidlink{0000-0002-3354-7294},\quad
    Mohamed Sallak\textsuperscript{1}\,\orcidlink{0000-0003-4803-2362},\quad
    Marc Ravaine\textsuperscript{2}\,\orcidlink{0009-0008-7939-1740}\par}
    \vspace{0.8em}
    {\small
    \textsuperscript{1}Université de technologie de Compiègne, CNRS, Alliance Sorbonne Université,\\ Heudiasyc, Compiègne, France\\
    \textsuperscript{2}ALTEN, Sophia Antipolis, France\\[0.3em]
    \texttt{\{jules.clerc,domitile.lourdeaux,mohamed.sallak\}@hds.utc.fr}\par}
\end{center}

\ifieeenotice
\begingroup
\let\thefootnote\relax
\footnotetext{\scriptsize \copyright~\copyrightYear{} IEEE. Personal use of this material is permitted. Permission from IEEE must be obtained for all other uses, in any current or future media, including reprinting/republishing this material for advertising or promotional purposes, creating new collective works, for resale or redistribution to servers or lists, or reuse of any copyrighted component of this work in other works.%
\ifx\vorCitation\empty\else{} This is the accepted version of: \vorCitation.\fi
\ifx\vorDOI\empty\else{} The final published version is available at \url{https://doi.org/\vorDOI}.\fi}
\endgroup
\fi

\vspace{0.6em}
\begin{center}
\begin{minipage}{0.9\textwidth}
\small
\noindent\textbf{Abstract.}
Interactive narratives let users shape a story while an Experience Manager (EM) keeps it coherent, balancing player freedom against authorial control. Large Language Models now enable a new generation of hybrid EMs, which combine run-time generation from free-text input with symbolic narrative control. However, their heterogeneity makes them hard to compare: A manager's behaviour is generally tied to the unique and specific scenario in which it operates, and much of its decision-making takes place outside of any explicit and verifiable structure. We address this with an LLM-based narrative platform whose managers are interchangeable modules operating over a shared symbolic world model of facts, operations, and goals. Because every applied operation carries explicit preconditions and effects, each session can be analysed as a plan through a common state-transition representation and evaluated, independently of how the manager works internally, with shared indicators of four properties: player agency, EM controllability, author intent, and robustness. We implement three contrasting managers, based respectively on open generation, Monte Carlo repair, and autonomous characters, and compare them through functional analysis on an interactive Little Red Riding Hood scenario. The results show that no manager dominates on every property, giving authors and EM designers a concrete basis to select a manager according to their priorities.

\end{minipage}
\end{center}
\vspace{1em}

\section{Introduction}







While traditional narratives are perceived passively, interactive narrative systems (INS) place the user at the centre of the plot, allowing them to influence the story through their choices. Preserving narrative coherence in the face of this freedom is the role of the Experience Manager (EM): a system-side agent that decides, at each stage, how far to follow the player and when to redirect the narrative. Reconciling player initiative with narrative coherence is a long-standing concern in the field~\cite{riedlInteractiveNarrativeIntelligent2013}, and every EM embodies one particular way of striking that balance.

The arrival of Large Language Models (LLMs) has been a turning point. Because such a model can interpret free-form textual input and generate plausible content as the story unfolds, the author no longer has to spell out in advance every action the player might attempt, nor every way the characters might react. This flexibility, however, comes at a price: when narrative decisions are made within the generative model, the EM can no longer easily anticipate them, keep them within the limits defined by the author, or put them back into the narrative space. A recent line of research attempts to reconcile flexibility with control through hybrid designs, in which the manager itself combines the fluidity of generation with the logic of an explicit structure~\cite{han-etal-2024-ibsen, senanayakeLanguageModelsNarrative2025, ngawAuthoringOpenEndedBehaviors2024}, for example, by letting a language model guide a planner, or by integrating the generated content into an inspectable game state.

The wide variety of these hybrid designs, however, makes comparison difficult. Since an EM is generally presented in a unique and personalised scenario, its qualities are inseparable from those of the scenario it runs on~\cite{magerkoComparativeAnalysisStory2007}: the field still lacks shared benchmarks, a situation known as the "No Yardstick" problem~\cite{koenitzChallengesIDNResearch2019}. Efforts to structure the domain have primarily focused on architecture, unifying how managers are described, how they are decomposed, and how they interact with their environments~\cite{thueGeneralizedExperienceManagement2015, moriStructuredAnalysisExperience2019, moriEMGluePlatformDecoupling2022}. They leave open the complementary question that we address here: how to judge, on common ground, the narratives that different EMs actually produce?

To investigate this question, we consider three contrasting forms of hybrid experience management, evaluated separately: open generation, targeted repair, and character-driven progression. Their different strategies lead us to three hypotheses. \textit{H1, the Monte Carlo Tree Search (MCTS) repair improves robustness}: We expect the MCTS-based manager to be more robust than the open manager, since it behaves identically during ordinary interaction but projects a corrective sequence when the narrative reaches a dead end. \textit{H2, the multi-agent manager increases emergent progression}: We expect the multi-agent manager to produce more emergent and less authored narratives, as autonomous NPCs generate and execute events independently of the player. \textit{H3, no architecture dominates}: We expect no manager to perform best on all four properties, since each architecture prioritises a different balance of player agency, EM controllability, author intent, and robustness.

Our solution involves decoupling the internal workings of an EM from how its behaviour is measured. We build a narrative platform whose managers are interchangeable modules connected to the same symbolic world model, expressed in terms of facts, operations, and goals. Every operation that is effectively applied carries its own preconditions and effects, so a session played under any manager can be replayed as a path through a formal state-transition representation. Based on this common foundation, the narratives are then compared using indicators of four properties (player agency, EM controllability, author intent, and robustness), covering the different actors in the system (author, EM, player). The paper offers (1) a modular LLM-based platform where heterogeneous hybrid managers share a common world model; (2) three contrasting managers: one favours open generation, another adds Monte Carlo repair, and the third delegates events to autonomous characters (with the ambition of placing them at different points on the spectrum of agency and control); and (3) a functional analysis that brings their sessions into a single formalism and reads them with shared indicators, demonstrated on an interactive \textit{Little Red Riding Hood} scenario.

\section{Related Work}
Experience management addresses a central problem in interactive narrative: the system must react to player actions while preserving narrative progression~\cite{riedlInteractiveNarrativeIntelligent2013}. Planning-based systems represent actions, their conditions, their effects, and narrative objectives, allowing control knowledge to be expressed as constraints over possible state trajectories~\cite{porteousApplyingPlanningInteractive2010}. Search-based drama management~\cite{nelsonSearchBasedDramaManagement2005} evaluates alternative interventions according to their expected narrative value, while plan-based mediation attempts to preserve the coherence of an executing narrative plan when the player performs an unanticipated action~\cite{youngArchitectureIntegratingPlanbased2004a}. Control can also be distributed among autonomous characters: character-centred systems derive narrative behaviour from the goals and decisions of individual agents~\cite{siThespianUsingMultiagent2005}, whereas multi-agent story-graph pruning constrains NPC actions according to their long-term consequences while leaving player actions available~\cite{wareMultiAgentNarrativeExperience2019}. MCTS provides another way to explore large narrative spaces by using believability estimates to guide the search towards stories satisfying user-defined goals~\cite{kartalUserdrivenNarrativeVariation2014}. These approaches differ in their decision mechanisms, but all rely on an explicit representation on which narrative control can operate.

LLMs challenge this assumption by allowing players and characters to act beyond a fully predefined set of dialogues or behaviours. Recent systems attempt to retain narrative direction through complementary mechanisms: director agents coordinate actor agents and reschedule the plot after player intervention~\cite{han-etal-2024-ibsen}; plot-based reflection aligns generated reactions with player intentions~\cite{wuEnhancedImmersionAgency2025a}; and hierarchical memory supports coherence and character behaviour over longer interactions~\cite{xuOpenTheatreOpenSourceToolkit2025}. These systems extend the range of interactions, but they also move part of narrative decision-making outside the explicit structures traditionally used by an EM.

Hybrid approaches reconnect these generative capabilities with symbolic or authored control mechanisms. A language model can guide a classical narrative-planning algorithm as a search heuristic~\cite{senanayakeLanguageModelsNarrative2025}, while behaviour-tree systems~\cite{ngawAuthoringOpenEndedBehaviors2024} can use generated dialogue and descriptions without relinquishing authored goals or explicit game-state updates. Our platform follows this direction but applies it to several EM strategies rather than to a single architecture. LLM components are used to interpret player commands, generate operations that modify the world state, propose events during narrative projection, and drive autonomous NPC decisions. At the same time, every operation effectively applied to the story is represented through explicit preconditions and effects. This shared symbolic layer allows for the introduction of openness to different techniques for managing interactive narrative without abandoning an inspectable description of world evolution.

The resulting architectures are nevertheless difficult to compare, and existing work has mainly addressed this fragmentation from an architectural or integration perspective. The Generalized Experience Management framework~\cite{thueGeneralizedExperienceManagement2015} provides a common representation of experience-management processes, while structured analysis~\cite{moriStructuredAnalysisExperience2019} identifies comparable components across managers with different designs. EM-Glue~\cite{moriEMGluePlatformDecoupling2022} addresses software interoperability by decoupling managers from their execution environments through an intermediate platform and communication protocol. These contributions help describe, decompose, or connect heterogeneous systems, but not evaluate the narrative trajectories these systems produce on common functional criteria, which is the gap we address.

\section{Narrative Platform Design}
\begin{figure}[t]
    \centering
    \includegraphics[width=0.62\linewidth]{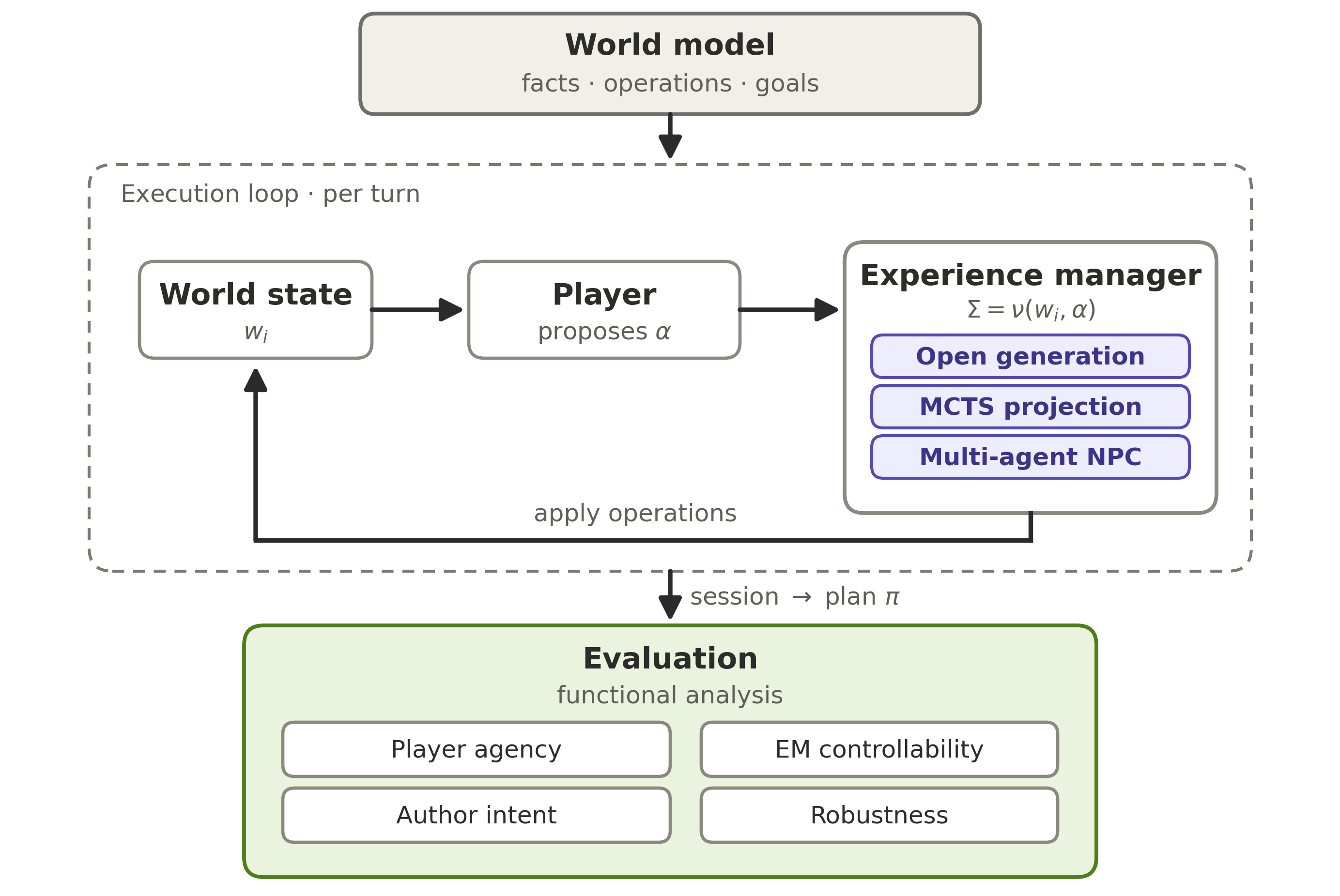}
    \caption{Overall architecture of the platform}
    \label{fig:architecture}
\end{figure}

Our platform has three layers (Fig.~\ref{fig:architecture}): a world model of facts, operations, and goals; an execution loop in which the player issues natural-language commands and a swappable EM selects the operations applied; and an evaluation layer that maps sessions onto a formal model (Section~\ref{sec:evaluation}).
 
\subsection{Symbolic world model}
\label{subsec:world-model}
We represent the narrative world with symbolic facts, in a way close to ontologies. A narrative fact is a subject-predicate-object triple:
\begin{equation}
\tau = (u, p, o),\qquad u \in \mathcal{U}, \quad p \in \mathcal{P}, \quad o \in \mathcal{U} \cup \mathcal{L}
\end{equation}

where $\mathcal{U}$ is the set of entities, $\mathcal{P}$ the set of predicates and $\mathcal{L}$ the set of literal values. The set of all possible facts is $\mathcal{F} \subseteq \mathcal{U} \times \mathcal{P} \times (\mathcal{U} \cup \mathcal{L})$. A world state is a snapshot of the facts held true at a given time, $w \subseteq \mathcal{F}$, and we write $W$ for the set of world states.
 
The world evolves through narrative operations. An operation $\eta = (\mathit{Pre}_\eta, \mathit{Add}_\eta, \mathit{Del}_\eta)$, with $\mathit{Pre}_\eta, \mathit{Add}_\eta, \mathit{Del}_\eta \subseteq \mathcal{F}$ represents a possible modification of the current world state. The preconditions $\mathit{Pre}_\eta$ specify the facts that must be true before the operation can be performed, while $\mathit{Add}_\eta$ and $\mathit{Del}_\eta$ respectively specify the facts that become true and the facts that cease to be true after its application. It is applicable in $w$ if $\mathit{Pre}_\eta \subseteq w$, and its application leads to the world state:
\begin{equation}
\mathit{Result}(w, \eta) = (w \setminus \mathit{Del}_\eta) \cup \mathit{Add}_\eta .
\end{equation}

We separate the operations the player may propose, the actions $\mathcal{A}$, from those the system controls, the events $\mathcal{E}$, and write $\mathcal{H} = \mathcal{A} \cup \mathcal{E}$ (with $\mathcal{A} \cap \mathcal{E} = \emptyset$) for the set of operations. The operations applicable in $w$, together with their restrictions to actions and events are:
\begin{equation}
    \begin{aligned}
        \mathcal{H}(w) &= \{\, \eta \in \mathcal{H} \mid \mathit{Pre}_\eta \subseteq w \,\}, \\
        \mathcal{A}(w) &= \mathcal{H}(w) \cap \mathcal{A}, \\
        \mathcal{E}(w) &= \mathcal{H}(w) \cap \mathcal{E}
    \end{aligned}
\end{equation}

The operation set $\mathcal{H}$ is not fixed. An EM may enlarge it at run time by generating new operations, as detailed in Section~\ref{sec:em-architectures}. Finally, a narrative goal is a set of facts $\mathcal{G} \subseteq \mathcal{F}$, and a world state is terminal when $\mathcal{G} \subseteq w$.
 
\subsection{Execution loop}
 
The interaction proceeds turn by turn. At each step, the player does not directly access the current world state $w \in W$. Instead, the platform provides a textual observation describing what is perceptible in the narrative world. The player may then answer with a natural-language command, or remain inactive. We denote by $\mathcal{C}$ the set of natural-language commands and by $\perp$ an inaction. The player's input at a given turn is:
\begin{equation}
    \alpha \in \mathcal{C} \cup \{\perp\}
\end{equation}

The input $\alpha$ does not directly modify the world state. It is only a proposal expressed by the player and must be resolved by the Manager. The EM selects the narrative operations effectively applied during the turn through a policy:
\begin{equation}
    \nu : W \times (\mathcal{C} \cup \{\perp\})
    \to \mathcal{H}^{*}
\end{equation}

Given the current world state $w$ and the player's input $\alpha$, the EM
selects a finite sequence of operations:
\begin{equation}
    \Sigma = \nu(w,\alpha)
\end{equation}

The selected sequence may contain an existing action matching the player's command, generated actions or events, system events, or no operation at all. The concrete way in which this sequence is produced depends on the EM architecture and is detailed in Section~\ref{sec:em-architectures}.

If a sequence $\Sigma = (\eta_1,\ldots,\eta_k)$ is selected, its operations are applied in order to the current world state. The resulting state is defined as:
\begin{equation}
    w' = \mathit{Result}^{*}(w,\Sigma),
\end{equation}
where $\mathit{Result}^{*}$ denotes the successive application of $\mathit{Result}$ to all operations in $\Sigma$. A complete execution produces a sequence of world states $(w_0,w_1,\ldots,w_n)$, which represents the narrative trajectory generated during the session. We write $(\,)$ for the empty sequence, with the convention $\mathit{Result}^{*}(w,(\,)) = w$ (no operation applied, the world state is left unchanged).

\section{Experience Manager Architectures}
\label{sec:em-architectures}
All EMs share two LLM-based functions that turn a command $c\in\mathcal{C}$ into a formal operation. They differ in how they use them and in the control they exert over the story.

\paragraph{Natural-language matching ($\delta$)}
The first maps a command to an existing applicable operation. Given a command $c$ and an operation type $k\in\{\textit{action},\textit{event}\}$,
\begin{equation}
\delta : \mathcal{C}\times\{\textit{action},\textit{event}\}\to \mathcal{H}\cup\{\beth\}
\end{equation}

returns the operation of type $k$ that matches $c$ when one exists, and the symbol $\beth$ when none does.

\paragraph{Dynamic generation ($\mathit{Gen}$)}
When matching fails, an EM may create a new operation. Let $\mathcal{H}_{all}$ be the set of all narrative operations that can be built over the world facts, of which the current operation set is a subset, $\mathcal{H} \subseteq \mathcal{H}_{all}$. The function
\begin{equation}
\mathit{Gen} : W \times \mathcal{C} \times \{\textit{action},\textit{event}\}\to \mathcal{H}_{all}
\end{equation}

returns a fresh operation $\eta_{new} \in \mathcal{H}_{all} \setminus \mathcal{H}$ of type $k$, synthesised from the command $c$ in the state $w$. The EM then adds it to the current set, $\mathcal{H}:=\mathcal{H}\cup\{\eta_{new}\}$, which is how the narrative space is enlarged at run time.

\subsection{Open Generation-Based EM}
This manager resolves one operation per turn and maximises openness to the player. It injects an event, drawn uniformly at random from $\mathcal{E}(w)$, when a draw $u\sim\mathcal{U}([0,1])$ falls below $p_{evt}$ or the player submits no command, provided an applicable event exists (lines~1--3); this pushes the story forward independently of the player. Otherwise the command $c$ is processed: it is bound through the parser $\delta$ to an existing action when possible (lines~7--8), and otherwise triggers generation, adding a new action to $\mathcal{H}$ (lines~10--12). An empty command with no applicable event yields a wait (lines~4--5). The selected operation is finally validated against its preconditions (lines~15--18): it is executed only if $\mathit{Pre}_\eta\subseteq w$, otherwise the turn is rejected and $w$ is left unchanged. Algorithm~\ref{alg:em_open} details $\nu^{\textsc{open}}$.
 
\begin{algorithm}
    \caption{Open Generation-Based EM policy $\nu^{\textsc{open}}(w,\alpha)$}
    \label{alg:em_open}
    \textbf{Input}: world state $w$, player command $\alpha \in \mathcal{C} \cup \{\perp\}$ (natural language, or $\perp$ for inaction)\\
    \textbf{Parameter}: event probability $p_{evt}$\\
    \textbf{Output}: effective operation sequence $\Sigma$
    
    \begin{algorithmic}[1]
        \STATE $u \sim \mathcal{U}([0,1])$
        \IF{$(u < p_{evt}$ \textbf{or} $\alpha = \perp)$ \textbf{and} $\mathcal{E}(w) \neq \emptyset$}
            \STATE $\eta \leftarrow \mathrm{Uniform}(\mathcal{E}(w))$ \COMMENT{event injection}
        \ELSIF{$\alpha = \perp$}
            \STATE \textbf{return} $(\,)$ \COMMENT{no event available: wait}
        \ELSE
            \IF{$\delta(\alpha,\textit{action}) \neq \beth$}
                \STATE $\eta \leftarrow \delta(\alpha,\textit{action})$ \COMMENT{matched action}
            \ELSE
                \STATE $\eta_{new} \leftarrow \mathrm{Gen}(w,\alpha,\textit{action})$ \COMMENT{create a new operation}
                \STATE $\mathcal{H} \leftarrow \mathcal{H} \cup \{\eta_{new}\}$ \COMMENT{add it to $\mathcal{H}$}
                \STATE $\eta \leftarrow \eta_{new}$ \COMMENT{apply generated action}
            \ENDIF
        \ENDIF
        \IF{$\mathit{Pre}_\eta \subseteq w$}
            \STATE \textbf{return} $(\eta)$ \COMMENT{single-operation sequence}
        \ELSE
            \STATE \textbf{return} $(\,)$ \COMMENT{reject, $w$ unchanged}
        \ENDIF
    \end{algorithmic}
\end{algorithm}
 
Because any unrecognised command is turned into a new operation, this strategy grants strong player agency at the cost of an unbounded expansion of the narrative space. The EM's own control reduces to the injection of events and to precondition validation.

\subsection{MCTS Projection-Based EM}
This manager extends the open one with a repair mechanism that improves robustness without giving up its openness. As long as at least one operation is applicable, it behaves exactly like $\nu^{\textsc{open}}$ (lines~1--2). The additional mechanism is triggered only when the system reaches a \emph{dead end}, defined as a state in which no operation is currently applicable:
\begin{equation}
\mathit{blocked}(w) \iff \mathcal{H}(w)=\emptyset .
\end{equation}

From such a state, rather than leaving the narrative stuck, the EM runs an MCTS over event continuations to project a sequence that brings the story back towards a finishable narrative space, and injects it in the system. Algorithm~\ref{alg:em_mcts} details the resulting policy $\nu^{\textsc{mcts}}$.

From the root state $w$, the search builds a tree whose edges are new events. To expand a node, the EM queries an LLM for up to $b$ natural-language suggestions of plausible next events in the node's state, then selects one and instantiates it through $\mathit{Gen}$ (lines~13--14). A node is fully expanded once its $b$ children have been generated (line~8). These generated events live only in a working copy used by the search, and do not alter the live operation set $\mathcal{H}$. They support the rollouts and the children of the node, but only the retained sequence is committed to the system at the end. The events composing a candidate continuation $\sigma=(\eta_1,\dots,\eta_k)$ are therefore all generated ($\eta_i\in \mathcal{H}_{all}\setminus \mathcal{H}$). During traversal, children are selected by the Upper Confidence bound applied to Trees (line~9), an unvisited child being given infinite priority:
\begin{equation}
\mathrm{UCT}(n')=\frac{R(n')}{N(n')}+c\,\sqrt{\frac{\ln N(n)}{N(n')}},
\end{equation}

where $N(\cdot)$ is a visit count, $R(n')$ the accumulated reward of child $n'$, $n$ its parent, and $c$ the exploration constant. Each node is then evaluated by a bounded number of depth-limited random rollouts (line~17), measuring whether the goal becomes reachable:
\begin{equation}
r(w_n)=
    \begin{cases}
        1 & \text{if some rollout reaches a } w' \text{ with } \mathcal{G}\subseteq w',\\
        0 & \text{otherwise.}
    \end{cases}
\end{equation}

After $N$ iterations, the retained sequence $\sigma^{*}(w)$ is read off greedily from the root (line~20), following the most promising children by visit count, accumulated reward, then solved status. Being all newly generated, its events are then committed to $\mathcal{H}$ and executed in order (lines~22--30).

\begin{algorithm}
    \caption{MCTS Projection-Based EM policy $\nu^{\textsc{mcts}}(w,\alpha)$}
    \label{alg:em_mcts}
    \textbf{Input}: world state $w$, player command $\alpha \in \mathcal{C} \cup \{\perp\}$\\
    \textbf{Parameter}: exploration constant $c$, number of iterations $N$, suggestions per node $b$, rollouts per node $B$, rollout depth $d_{max}$\\
    \textbf{Output}: effective operation sequence $\Sigma$
    
    \begin{algorithmic}[1]
        \IF{$\mathcal{H}(w) \neq \emptyset$}
            \STATE \textbf{return} $\nu^{\textsc{open}}(w,\alpha)$ \COMMENT{not blocked: ordinary policy}
        \ENDIF
        \STATE $root \leftarrow \mathrm{Node}(w)$
        \FOR{$i \leftarrow 1$ \TO $N$}
            \STATE $n \leftarrow root$
            \STATE $w_n \leftarrow w$ \COMMENT{node state, starts at the blocked root}
            \WHILE{$n$ has $b$ children \textbf{and} $n$ is not terminal}
                \STATE $n \leftarrow \arg\max_{n' \in \mathit{children}(n)} \mathrm{UCT}(n')$
                \STATE $w_n \leftarrow \mathit{Result}(w_n, \eta_n)$
            \ENDWHILE
            \IF{$n$ is not terminal}
                \STATE $\eta \leftarrow \textsc{GenerateEvent}(w_n)$ \COMMENT{LLM suggestion instantiated via Gen}
                \STATE $n \leftarrow \textsc{AddChild}(n, \eta)$
                \STATE $w_n \leftarrow \mathit{Result}(w_n, \eta)$
            \ENDIF
            \STATE $r \leftarrow \textsc{Rollout}(w_n, B, d_{max})$
            \STATE $\textsc{Backpropagate}(n, r)$
        \ENDFOR
        \STATE $\sigma^{*}(w) \leftarrow \textsc{BestEventSequence}(root)$
        \STATE $\Sigma \leftarrow (\,)$
        \FORALL{$\eta_i \in \sigma^{*}(w)$}
            \STATE $\mathcal{H} \leftarrow \mathcal{H} \cup \{\eta_i\}$ \COMMENT{commit generated event to the live system}
            \IF{$\mathit{Pre}_{\eta_i} \subseteq w$}
                \STATE $w \leftarrow \mathit{Result}(w, \eta_i)$
                \STATE $\Sigma \leftarrow \Sigma \cdot (\eta_i)$
            \ELSE
                \STATE \textbf{break} \COMMENT{stop injection on first failure}
            \ENDIF
        \ENDFOR
        \STATE \textbf{return} $\Sigma$
    \end{algorithmic}
\end{algorithm}

This projection-based repair thus takes authority over the event channel only in blocked states: its control is stronger than that of $\nu^{\textsc{open}}$, yet local and corrective, and it trades player agency in those states for finishability.

\subsection{Multi-Agent NPC-Based EM}
This manager decentralises the system side into a set of autonomous non-player characters (NPCs) $N=\{c_1,\dots,c_m\}$, turning narrative progression into emergent, character-driven behaviour. A turn unfolds in two phases:
\begin{itemize}
    \item The \textit{player phase} (lines~2--5) applies the open policy with $p_{evt}=0$ when the player proposes a command $\alpha \neq \perp$. The command is matched or generated as in the open manager, but no event is injected on a probability draw.
    \item The \textit{NPC phase} then lets each character act once, in a random order $\rho$ drawn afresh every turn to avoid any first-mover advantage (lines~6--27).
\end{itemize}
 
Each character $c$ is driven by an LLM initialised with its profile and motivation $\theta_c\in\Theta_c$, where $\Theta_c$ is the space of profiles and motivations (persona, goals, traits) of character $c$. It acts only during the NPC phase: it triggers an applicable event, proposes a new one described by a natural-language text $t \in \mathcal{C}$, or waits,
\begin{equation}
    \nu_c : W \times \Theta_c \to \{\,\textsc{perform\_event}(\eta),\ \textsc{propose}(t),\ \textsc{wait}\,\}.
\end{equation}

Its decision $d_c=\nu_c(w,\theta_c)$ is then resolved into an effective operation in the current state (lines~12--22): a performed or proposed event is committed only when it is applicable in the current state ($\mathit{Pre}\subseteq w$), the generated case being added to $\mathcal{H}$ beforehand, otherwise the character waits and no operation is applied. The state is updated after each character acts (lines~23--26), until the narrative goal is reached (lines~8--9). Algorithm~\ref{alg:em_ma} details the resulting policy $\nu^{\textsc{MA}}$.

\begin{algorithm}
    \caption{Multi-Agent NPC-Based EM policy $\nu^{\textsc{ma}}(w,\alpha)$}
    \label{alg:em_ma}
    \textbf{Input}: world state $w$, player command $\alpha \in \mathcal{C} \cup \{\perp\}$\\
    \textbf{Parameter}: NPC set $N=\{c_1,\dots,c_m\}$ with profiles $\{\theta_c\}$\\
    \textbf{Output}: effective operation sequence $\Sigma$
    
    \begin{algorithmic}[1]
        \STATE $\Sigma \leftarrow (\,)$
        \IF{$\alpha \neq \perp$}
            \STATE $\Sigma \leftarrow \nu^{\textsc{open}}(w,\alpha)$ \COMMENT{player phase: only if the player proposes something, with $p_{evt}=0$}
            \STATE $w \leftarrow \mathit{Result}^{*}(w,\Sigma)$
        \ENDIF
        \STATE $\rho \leftarrow$ random permutation of $N$ \COMMENT{avoid first-mover advantage}
        \FOR{$j \leftarrow 1$ \TO $m$}
            \IF{$\mathcal{G} \subseteq w$}
                \STATE \textbf{break} \COMMENT{goal reached}
            \ENDIF
            \STATE $c \leftarrow \rho(j)$ \COMMENT{select NPC} \label{open:test}
            \STATE $d_c \leftarrow \nu_c(w,\theta_c)$ \COMMENT{LLM agent decision}
            \STATE $\eta^{c} \leftarrow \beth$ \COMMENT{no operation by default}
            \IF{$d_c = \textsc{perform\_event}(\eta)$ \textbf{and} $\eta \in \mathcal{E}(w)$}
                \STATE $\eta^{c} \leftarrow \eta$ \COMMENT{existing applicable event}
            \ELSIF{$d_c = \textsc{propose}(t)$}
                \STATE $\eta \leftarrow \mathrm{Gen}(w,t,\textit{event})$ \COMMENT{candidate event}
                \IF{$\mathit{Pre}_{\eta} \subseteq w$}
                    \STATE $\mathcal{H} \leftarrow \mathcal{H} \cup \{\eta\}$ \COMMENT{playable: commit}
                    \STATE $\eta^{c} \leftarrow \eta$
                \ENDIF
            \ENDIF
            \IF{$\eta^{c} \neq \beth$}
                \STATE $w \leftarrow \mathit{Result}(w,\eta^{c})$
                \STATE $\Sigma \leftarrow \Sigma \cdot (\eta^{c})$
            \ENDIF
        \ENDFOR
        \STATE \textbf{return} $\Sigma$
    \end{algorithmic}
\end{algorithm}
 
By partitioning the event channel $E$ among autonomous agents, this manager makes the narrative progress largely independently of the player, allowing a high event rate and through guarded NPC generation, emergent off-script behaviour kept consistent by the playability check. It is the most decentralised of the three: authority over events is distributed across the characters rather than held by a single policy, trading tight authorial control for emergent, character-driven narration.

\section{Evaluation Framework}
\label{sec:evaluation}
We evaluate the managers by functional analysis: each drives many sessions, recorded as plans in the state-transition representation of \cite{clercModelingInteractiveNarrative} and scored with properties defined on it.

\subsection{Simulation protocol}
Since the platform is LLM-based and the player interacts in natural language, we replace the human with an LLM\footnote{We use GPT-5 mini, with the temperature set to 1, both for the simulated player and for the LLM-based mechanisms of the Experience Managers.} acting as the player. At each turn it reads the textual observation and either proposes a single free-text command or does nothing ($\alpha\in\mathcal{C}\cup\{\bot\}$). Executing a simulation turns one session into a narrative trajectory $(w_0,\dots,w_n)$ with the operations effectively applied. A session ends when the goal is reached ($\mathcal{G}\subseteq w$), when no operation can advance the story, or when a turn budget is exhausted.

These simulations only approximate human behaviour, but they give EM designers fast feedback on their managers' capabilities, well before any costly user study. For each manager we run $n_{sim}=100$ sessions on a fixed scenario. We fixed this number by a convergence criterion, adding sessions until the running average of the properties, all valued in $[0,1]$, varies by less than $\varepsilon=0.005$ between successive sessions. This is reached at around one hundred sessions per manager, which we round to $n_{sim}$.

All three managers ($\nu^{\textsc{open}}$, $\nu^{\textsc{mcts}}$, $\nu^{\textsc{ma}}$) are evaluated on the same fixed scenario, a story inspired by Little Red Riding Hood that we reuse from earlier work on the formal modelling of INS. The player is the hunter, who must intervene at the right moments to keep the wolf from harming Little Red Riding Hood and her grandmother. Together with the wolf, these characters act as non-player characters. We chose it because it exposes the tension we study: some player choices derail the intended progression; killing the wolf too soon, for instance, blocks the events that should still occur and strands the story in a dead end, while its key stages are encoded as islands that a session is expected to cross.

\subsection{From sessions to plans}
To analyse all sessions on common ground (and subsequently compare managers with heterogeneous designs), we convert the world model design into an extended state-transition system \cite{clercModelingInteractiveNarrative} $\mathfrak{S}=(S,T,\gamma,s_{\mathrm{init}},S_{\mathrm{goal}})$: each world state $w_i$ is read as a state $s_i$, each operation in $\mathcal{H}=\mathcal{A}\cup\mathcal{E}$ as a transition of the same type, the $\mathit{Result}$ function as the transition function $\gamma$, and the goal facts $\mathcal{G}$ as the goal states $S_{\mathrm{goal}}=\{s\mid\mathcal{G}\subseteq s\}$. Mapped this way, the trajectory of one session becomes a path in $\mathfrak{S}$, which we call a plan $\pi$. We write $\Pi_m$ for the set of plans collected for manager $m$.

\subsection{Evaluated properties}
Four properties, covering the player, the manager, and the author, are scored on each plan:
\begin{itemize}
  \item \emph{Player agency} captures the player's ability to take meaningful actions and observe their impact on the story. We measure it through the \emph{acceptance rate} (the share of player proposals the manager applies unchanged).
  \item \emph{EM controllability} captures how much the manager can orient or constrain the unfolding narrative. We measure it through the \emph{event rate} (the share of executed transitions that are system events) and the \emph{pre-emption rate} (the share of player proposals overridden by an event).
  \item \emph{Author intent} captures how much of the authored narrative structure is preserved. We measure it through the \emph{islands rate} (the share of authored key waypoints, called islands, the plan crosses) and the \emph{scripted rate} (the share of executed transitions that come from the authored design rather than being generated at run time).
  \item \emph{Robustness} captures the system's ability to keep the story finishable despite the player's behaviour. We measure it through the \emph{completion rate} (the share of plans reaching a goal state) and the \emph{execution rate} (the share of turns whose selected operation is applied without rejection).
\end{itemize}
Each property is computed on every plan, then summarised over $\Pi_m$ by its mean and standard deviation.

\section{Results}
We confront H1–H3 with the properties reported in Table~\ref{tab:results}. Dispersion is largest for author intent and robustness (standard deviations up to 0.41), since a session tends to succeed or fail as a whole, and smallest for the controllability of $\nu^{\textsc{ma}}$ (0.06), whose NPCs act at every turn.

\paragraph{H1 --- The MCTS repair improves robustness} The results confirm it: robustness rises from $0.58$ to $0.76$ and the completion rate from $0.52$ to $0.80$, while author intent ($0.47$$\to$$0.44$) barely moves. The repair only triggers in dead ends, leaving ordinary turns identical to the open manager. The injected repair events do raise controllability, though ($0.28$$\to$$0.40$). It does, however, cost player agency ($0.44$$\to$$0.27$): in blocked states the manager takes over the event channel, reducing the share of player-driven turns.

\paragraph{H2 --- The multi-agent manager increases emergent progression}
We expected $\nu^{\textsc{ma}}$ to produce more emergent, less authored narratives. This is supported: it reaches the highest controllability ($0.52$ against $0.28$--$0.40$), as each NPC applies events every turn, and the lowest author intent ($0.04$), as off-script character behaviour drifts away from the authored islands. This value also reflects a strict calibration: both indicators only count the points of passage and transitions created by the author, so a generated event diminishes the author's intention even if it fits the narrative. One prediction is contradicted, though: rather than curbing the player, the manager keeps the highest agency ($0.68$), because the player still acts every turn while NPC events are added on top instead of replacing player proposals\footnote{This is also due to the fact that we only analyse whether the player's actions are accepted, but do not capture the real impact on the state of the world and the player's feelings.}. Here the tension is between emergence and author intent, not between control and agency.

\paragraph{H3 --- No architecture dominates}
The simulations confirm it: $\nu^{\textsc{mcts}}$ leads on robustness but has the lowest agency, $\nu^{\textsc{ma}}$ leads on agency and control but collapses on author intent, and $\nu^{\textsc{open}}$ stays balanced, on par with $\nu^{\textsc{mcts}}$ on author intent. The three profiles are thus not mutually dominant.

\begin{table}[htbp]
    \caption{Measured properties on the Little Red Riding Hood scenario ($n_{sim}=100$ sessions per manager): mean $\pm$ standard deviation.}
    \begin{center}
        \small
        \setlength{\tabcolsep}{4pt}
        \renewcommand{\arraystretch}{1.3}
        \begin{tabular}{|l|c|c|c|}
            \hline
            \textbf{Property} & $\nu^{\textsc{open}}$ & $\nu^{\textsc{mcts}}$ & $\nu^{\textsc{ma}}$ \\
            \hline
            Player agency & $0.44 \pm 0.18$ & $0.27 \pm 0.14$ & $\mathbf{0.68} \pm 0.29$ \\
            \hline
            EM controllability & $0.28 \pm 0.20$ & $0.40 \pm 0.20$ & $\mathbf{0.52} \pm 0.06$ \\
            \hline
            Author intent & $\mathbf{0.47} \pm 0.41$ & $0.44 \pm 0.41$ & $0.04 \pm 0.16$ \\
            \hline
            Robustness & $0.58 \pm 0.40$ & $\mathbf{0.76} \pm 0.31$ & $0.63 \pm 0.39$ \\
            \hline
        \end{tabular}
        \label{tab:results}
    \end{center}
\end{table}

These results suggest that no EM is ideal. The optimal choice depends on the priorities of the experience, whether it is maintaining player control, ensuring the story's completion, or allowing characters to influence emerging situations. By translating these design goals into measurable indicators and properties, this framework offers story authors and EM designers a concrete basis for selecting an EM.

\paragraph{Limitations and scalability}
The results still need to be confirmed with human players, notably to validate that the proposed indicators reflect perceived experience. Moreover, a single scenario isolates the effect of managers without establishing generalisability. Regarding scalability, the properties are computed on the simulated plans rather than on the whole state space, so the evaluation cost grows linearly with the number and length of sessions.

\section{Conclusion}
Our goal was to make LLM-based hybrid experience managers comparable despite their architectural diversity. To achieve this, we designed a narrative platform where managers are interchangeable modules sharing a single symbolic model, and we expressed each operation actually applied through explicit preconditions and effects. This allows us to represent any played session, regardless of the manager that produced it, as a path through a common formal representation of state transition, and to read it with the same properties of player agency, EM controllability, author intent, and robustness. Three managers of contrasting design (open generation, Monte Carlo repair, and autonomous characters) were instantiated on this common framework and compared by functional analysis on a Little Red Riding Hood scenario. The comparison confirmed that each manager favours different properties, with none prevailing on all of them, turning the choice of a manager into an explicit, priority-driven trade-off rather than a search for a single optimal design.

This study was deliberately scoped to isolate and validate the evaluation mechanism itself. Players were simulated by a language model, a choice that gives EM designers quick and repeatable feedback well before a costly user study, and managers were compared on a single controlled scenario so that observed differences could be attributed to the managers rather than to the story. Even within this scope, the framework already makes heterogeneous, LLM-driven managers measurable and comparable on common ground. Building on these foundations, several directions extend the work. The simulated player can be complemented by human participants to confirm the indicators against perceived experience; the evaluation can be extended to richer and more varied scenarios; the properties can be deepened; and, since the managers share the same world model, they could be combined, for instance by selecting the most suitable one at each step. Expanding the catalogue of managers and properties would transform the platform into a shared testbed where heterogeneous experience management strategies would be designed, optimised and compared: an evaluation resource for the scientific community.

\begingroup
\small
\bibliographystyle{IEEEtran}
\bibliography{references}

\begin{thebibliography}{10}
\providecommand{\url}[1]{#1}
\csname url@samestyle\endcsname
\providecommand{\newblock}{\relax}
\providecommand{\bibinfo}[2]{#2}
\providecommand{\BIBentrySTDinterwordspacing}{\spaceskip=0pt\relax}
\providecommand{\BIBentryALTinterwordstretchfactor}{4}
\providecommand{\BIBentryALTinterwordspacing}{\spaceskip=\fontdimen2\font plus
\BIBentryALTinterwordstretchfactor\fontdimen3\font minus
  \fontdimen4\font\relax}
\providecommand{\BIBforeignlanguage}[2]{{%
\expandafter\ifx\csname l@#1\endcsname\relax
\typeout{** WARNING: IEEEtran.bst: No hyphenation pattern has been}%
\typeout{** loaded for the language `#1'. Using the pattern for}%
\typeout{** the default language instead.}%
\else
\language=\csname l@#1\endcsname
\fi
#2}}
\providecommand{\BIBdecl}{\relax}
\BIBdecl

\bibitem{riedlInteractiveNarrativeIntelligent2013}
M.~O. Riedl and V.~Bulitko, ``Interactive {{Narrative}}: {{An Intelligent
  Systems Approach}},'' \emph{AI Magazine}, vol.~34, no.~1, pp. 67--67, 2013.

\bibitem{han-etal-2024-ibsen}
\BIBentryALTinterwordspacing
S.~Han, L.~Chen, L.-M. Lin, Z.~Xu, and K.~Yu, ``{IBSEN}: Director-actor agent
  collaboration for controllable and interactive drama script generation,'' in
  \emph{Proceedings of the 62nd Annual Meeting of the Association for
  Computational Linguistics (Volume 1: Long Papers)}, L.-W. Ku, A.~Martins, and
  V.~Srikumar, Eds.\hskip 1em plus 0.5em minus 0.4em\relax Bangkok, Thailand:
  Association for Computational Linguistics, Aug. 2024, pp. 1607--1619.
  [Online]. Available: \url{https://aclanthology.org/2024.acl-long.88/}
\BIBentrySTDinterwordspacing

\bibitem{senanayakeLanguageModelsNarrative2025}
L.~Senanayake and S.~G. Ware, ``Language {{Models}} as {{Narrative Planning
  Heuristics}},'' in \emph{Proceedings of the 20th {{International Conference}}
  on the {{Foundations}} of {{Digital Games}}}, ser. {{FDG}} '25.\hskip 1em
  plus 0.5em minus 0.4em\relax New York, NY, USA: Association for Computing
  Machinery, May 2025, pp. 1--9.

\bibitem{ngawAuthoringOpenEndedBehaviors2024}
B.~Ngaw, G.~Jena, J.~Sedoc, and A.~Normoyle, ``Towards {{Authoring Open-Ended
  Behaviors}} for {{Narrative Puzzle Games}} with {{Large Language Model
  Support}},'' in \emph{Proceedings of the 19th {{International Conference}} on
  the {{Foundations}} of {{Digital Games}}}.\hskip 1em plus 0.5em minus
  0.4em\relax Worcester MA USA: ACM, May 2024, pp. 1--4.

\bibitem{magerkoComparativeAnalysisStory2007}
B.~Magerko, ``A {{Comparative Analysis}} of {{Story Representations}} for
  {{Interactive Narrative Systems}},'' \emph{Proceedings of the AAAI Conference
  on Artificial Intelligence and Interactive Digital Entertainment}, vol.~3,
  no.~1, pp. 91--94, 2007.

\bibitem{koenitzChallengesIDNResearch2019}
H.~Koenitz and M.~P. Eladhari, ``Challenges of {{IDN Research}} and
  {{Teaching}},'' in \emph{Interactive {{Storytelling}}}, R.~E.
  {Cardona-Rivera}, A.~Sullivan, and R.~M. Young, Eds.\hskip 1em plus 0.5em
  minus 0.4em\relax Cham: Springer International Publishing, 2019, pp. 26--39.

\bibitem{thueGeneralizedExperienceManagement2015}
D.~J. Thue, ``Generalized {{Experience Management}},'' \emph{Doctoral Thesis},
  2015.

\bibitem{moriStructuredAnalysisExperience2019}
G.~Mori, D.~Thue, and S.~Schiffel, ``A structured analysis of experience
  management techniques,'' in \emph{Proceedings of the {{Fifteenth AAAI
  Conference}} on {{Artificial Intelligence}} and {{Interactive Digital
  Entertainment}}}, ser. {{AIIDE}}'19, vol.~15.\hskip 1em plus 0.5em minus
  0.4em\relax Atlanta, Georgia: AAAI Press, Oct. 2019, pp. 174--180.

\bibitem{moriEMGluePlatformDecoupling2022}
------, ``{{EM-Glue}}: {{A Platform}} for {{Decoupling Experience Managers}}
  and {{Environments}},'' \emph{Proceedings of the AAAI Conference on
  Artificial Intelligence and Interactive Digital Entertainment}, vol.~18,
  no.~1, pp. 266--274, Oct. 2022.

\bibitem{porteousApplyingPlanningInteractive2010}
J.~Porteous, M.~Cavazza, and F.~Charles, ``Applying {{Planning}} to
  {{Interactive Storytelling}}: {{Narrative Control Using State
  Constraints}},'' \emph{ACM TIST}, vol.~1, p.~10, Nov. 2010.

\bibitem{nelsonSearchBasedDramaManagement2005}
M.~Nelson and M.~Mateas, ``Search-{{Based Drama Management}} in the
  {{Interactive Fiction Anchorhead}},'' \emph{Proceedings of the AAAI
  Conference on Artificial Intelligence and Interactive Digital Entertainment},
  vol.~1, no.~1, pp. 99--104, Jun. 2005.

\bibitem{youngArchitectureIntegratingPlanbased2004a}
R.~Young, M.~Riedl, M.~Branly, A.~Jhala, {\relax RJ}.~Martin, and {\relax
  CJ}.~Saretto, ``An architecture for integrating plan-based behavior
  generation with interactive game environments,'' \emph{Journal of Game
  Development}, vol.~1, pp. 51--70, Jan. 2004.

\bibitem{siThespianUsingMultiagent2005}
M.~Si, S.~C. Marsella, and D.~V. Pynadath, ``Thespian: Using multi-agent
  fitting to craft interactive drama,'' in \emph{Proceedings of the Fourth
  International Joint Conference on {{Autonomous}} Agents and Multiagent
  Systems}, ser. {{AAMAS}} '05.\hskip 1em plus 0.5em minus 0.4em\relax New
  York, NY, USA: Association for Computing Machinery, Jul. 2005, pp. 21--28.

\bibitem{wareMultiAgentNarrativeExperience2019}
S.~G. Ware, E.~T. Garcia, A.~Shirvani, and R.~Farrell, ``Multi-{{Agent
  Narrative Experience Management}} as {{Story Graph Pruning}},''
  \emph{Proceedings of the AAAI Conference on Artificial Intelligence and
  Interactive Digital Entertainment}, vol.~15, no.~1, pp. 87--93, Oct. 2019.

\bibitem{kartalUserdrivenNarrativeVariation2014}
B.~Kartal, J.~Koenig, and S.~J. Guy, ``User-driven narrative variation in large
  story domains using monte carlo tree search,'' in \emph{Proceedings of the
  2014 International Conference on {{Autonomous}} Agents and Multi-Agent
  Systems}, ser. {{AAMAS}} '14.\hskip 1em plus 0.5em minus 0.4em\relax
  Richland, SC: {International Foundation for Autonomous Agents and Multiagent
  Systems}, May 2014, pp. 69--76.

\bibitem{wuEnhancedImmersionAgency2025a}
H.~Wu, W.~Wu, T.~Xu, J.~Zhang, and H.~Zhao, ``Towards {{Enhanced Immersion}}
  and {{Agency}} for {{LLM-based Interactive Drama}},'' in \emph{Proceedings of
  the 63rd {{Annual Meeting}} of the {{Association}} for {{Computational
  Linguistics}} ({{Volume}} 1: {{Long Papers}})}, W.~Che, J.~Nabende,
  E.~Shutova, and M.~T. Pilehvar, Eds.\hskip 1em plus 0.5em minus 0.4em\relax
  Vienna, Austria: Association for Computational Linguistics, Jul. 2025, pp.
  11\,166--11\,182.

\bibitem{xuOpenTheatreOpenSourceToolkit2025}
T.~Xu, H.~Wu, W.~Wu, and H.~Zhao, ``Open-{{Theatre}}: {{An Open-Source
  Toolkit}} for {{LLM-based Interactive Drama}},'' in \emph{Proceedings of the
  2025 {{Conference}} on {{Empirical Methods}} in {{Natural Language
  Processing}}: {{System Demonstrations}}}, I.~Habernal, P.~Schulam, and
  J.~Tiedemann, Eds.\hskip 1em plus 0.5em minus 0.4em\relax Suzhou, China:
  Association for Computational Linguistics, Nov. 2025, pp. 453--460.

\bibitem{clercModelingInteractiveNarrative}
\BIBentryALTinterwordspacing
J.~Clerc, D.~Lourdeaux, M.~Sallak, J.~Barbier, and M.~Ravaine, ``Modeling
  {{Interactive Narrative Systems}}: {{A Formal Approach}},'' in
  \emph{Proceedings of the 8th International Workshop on Computational Models
  of Narrative ({{CMN}} 2025)}, ser. {{CEUR}} Workshop Proceedings, vol. 4288,
  2025, pp. 179--187. [Online]. Available:
  \url{https://ceur-ws.org/Vol-4288/paper14.pdf}
\BIBentrySTDinterwordspacing

\end{thebibliography}
\endgroup

\end{document}